\documentclass{article}
\usepackage{spconf,amsmath,graphicx,amsfonts,booktabs,subcaption}
\usepackage[separate-uncertainty=true, multi-part-units=single]{siunitx}

\title{Text-to-image synthesis method evaluation based on visual patterns}
\name{William Lund Sommer and Alexandros Iosifidis} 
\address{Department of Engineering, ECE, Aarhus University, Denmark\\
E-mails: 201303715@post.au.dk, alexandros.iosifidis@eng.au.dk}
\begin{document}
%
\maketitle
\begin{abstract}
A commonly used evaluation metric for text-to-image synthesis is the Inception score (IS) \cite{inceptionscore}, which has been shown to be a quality metric that correlates well with human judgment. However, IS does not reveal properties of the generated images indicating the ability of a text-to-image synthesis method to correctly convey semantics of the input text descriptions. In this paper, we introduce an evaluation metric and a visual evaluation method allowing for the simultaneous estimation of the realism, variety and semantic accuracy of generated images. The proposed method uses a pre-trained Inception network \cite{inceptionnet} to produce high dimensional representations for both real and generated images. These image representations are then visualized in a $2$-dimensional feature space defined by the t-distributed Stochastic Neighbor Embedding (t-SNE) \cite{tsne}. Visual concepts are determined by clustering the real image representations, and are subsequently used to evaluate the similarity of the generated images to the real ones by classifying them to the closest visual concept. The resulting classification accuracy is shown to be a effective gauge for the semantic accuracy of text-to-image synthesis methods.
\end{abstract}
\section{Introduction}
\label{sec:intro}
The progress of text-to-image synthesis field has traditionally been empirically driven, and since no ``one-fits-all'' metric has been proposed for both quantitative and qualitative evaluation of machine-generated images, the existence of a robust evaluation metric is crucial for the advancement of the field. The Inception Score (IS) \cite{inceptionscore} is the commonly chosen metric for evaluating the quality of the generated images, as it has been shown to correlate well with human judgment. However, this metric solely takes into account the realism and variety of the generated images \cite{attngan,mirrorgan,hdgan,stackgan}. This works well for unrestricted image generation, but fails to capture important information when the image synthesis task is conditioned on classes or text descriptions. For conditional text-to-image synthesis, a suitable evaluation metric should also take into account how well the generated images correspond to the given conditions. For text-to-image synthesis methods this means the method's ability to correctly capture the semantic meaning of the input text descriptions. Human rankings give an excellent estimate of semantic accuracy but evaluating thousands of images following this approach is impractical, since it is a time consuming, tedious and expensive process.

In this paper, we propose a new evaluation method for text-to-image synthesis that allows for the simultaneous estimation of realism, variety and semantic accuracy of the generated images. We use t-Distributed Stochastic Neighbor Embedding (t-SNE) \cite{tsne} to visualize Inception-net \cite{inceptionnet} features of the real and generated images. Visual concepts are determined by clustering the representations of the real images in the low-dimensional feature space and are used to access their class-wise similarities. We evaluate four state-of-the-art text-to-image synthesis methods using this method, i.e. the AttnGAN \cite{attngan}, the HDGAN \cite{hdgan}, the MirrorGAN \cite{mirrorgan}, and the StackGAN++ \cite{stackgan++}, and we show that the proposed evaluation method can capture information related to the classes of the corresponding real (ground-truth) images.

\section{Proposed Evaluation Method}
An ideal evaluation method for text-to-image synthesis methods should be able to highlight several key properties of the generated images. The first property is realism and image quality, i.e. to what extend the generated images resemble real images. The second property is variety, i.e., the generated images should ideally show as much variety as that appearing in real images. Moreover, mode collapse, i.e. different captions leading to the same image being generated multiple times, should be avoided. The third key property is the methods' ability to accurately capture the semantics of the provided text descriptions. The evaluation method introduced in this paper focuses on all of these three properties. The procedure followed by the proposed method is described next.

\begin{figure*}
	\centering
		\includegraphics[width=0.8\linewidth]{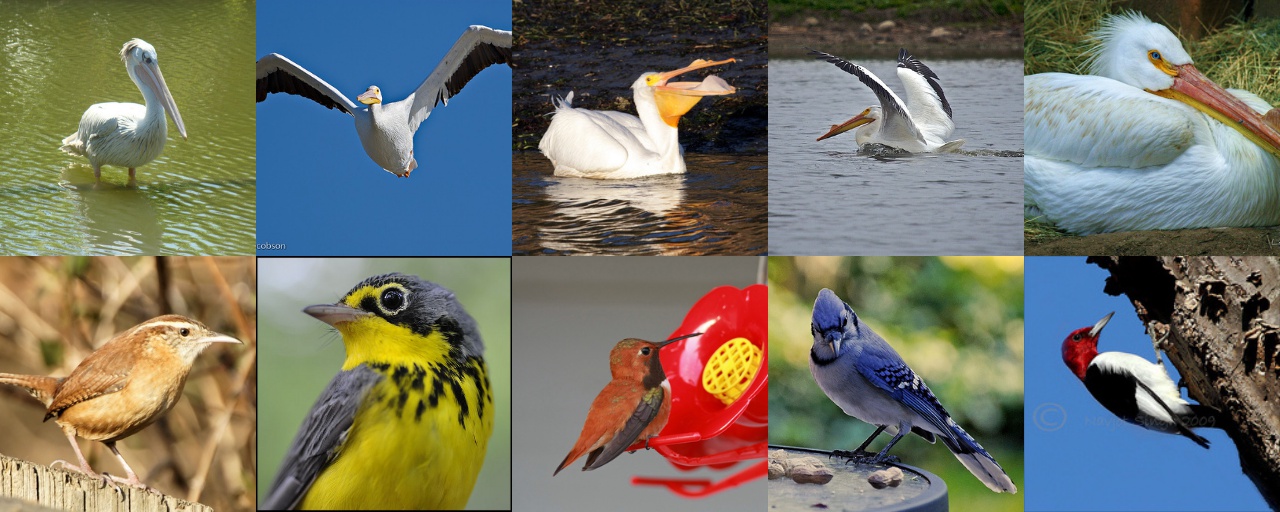}
		\label{fig:attngan}
	\caption{Example real images from the CUB-200-2011 data set \cite{cub200}. Upper row: Examples from the White Pelican class. Lower row: Examples from the Carolina Wren, Canada Warbler, Rufous Hummingbird, Blue Jay, and Red Headed Woodpecker classes in the order given.}
	\label{fig:CUB_images}
\end{figure*}

Let us denote by $X=\left\lbrace x_1, x_2, \dots , x_n\right\rbrace$ a set of $n$ real images, each followed by a text description (in the form of a caption) included in the set $T = \left\lbrace t_1, t_2, \dots ,t_n\right\rbrace$. A text-to-image synthesis method, usually based on a generative network $G$, generates synthetic images included in the set $S =  \left\lbrace s_1, s_2,\dots s_n\right\rbrace$, where $s_i = G(t_i)$ for $i=1,2,\dots,n$. As commonly done in the evaluation of text-to-image synthesis methods, the data sets $(X,T)$ should not have been used for training of $G$, and the classes of $(X,T)$ should be disjoint from the classes of the data used for training $G$\footnote{In general there may be several captions for each real image, and thus more synthetic images than the real ones. Here one caption per image is assumed to simplify notation.}.

To capture semantic information related to the classes of the real images in the set $X$, an Inception network \cite{inceptionnet} is trained to achieve high classification performance ($\sim 95\%$) on the real images $X$. This network is subsequently used to extract image representations for both the real images in $X$ and the synthetic images in $S$. This is done by introducing the images to the network and extracting the next to the last layer outputs. Let us denote by 
$\tilde{X} = \left\lbrace \tilde{x}_1, \tilde{x}_2, \dots, \tilde{x}_n \right\rbrace$ and $\tilde{S} = \left\lbrace \tilde{s}_1, \tilde{s}_2, \dots, \tilde{s}_n \right\rbrace$, where $\tilde{x}_i \in \mathbb{R}^D$ and $\tilde{s}_i \in \mathbb{R}^D, \:i=1,\dots,n$, the representations of the real and synthetic images, respectively. Since the next to the last layer in Inception network is formed by $1024$ neurons, we have $D = 1024$.

To obtain $2$-dimensional representations of the real and synthetic images, the two sets ($\tilde{X}, \tilde{S}$) are combined and used for applying the t-SNE \cite{tsne} mapping. In our experiments we followed the process suggested by the authors in \cite{tsne}, i.e. an intermediate dimensionality reduction from $D$ to $50$ dimensions is applied by using Principal Component Analysis (PCA) \cite{pca}. When performing the t-SNE mapping, the perplexity is set to the number of images in each class since this is the expected number of neighbors. Let us denote by $\Phi^X \in \mathbb{R}^{2 \times n}$ and $\Phi^S \in \mathbb{R}^{2 \times n}$ the resulting $2$-dimensional representations of the real and the synthetic images, respectively. 

\begin{figure}[ht]
	\centering
	\begin{subfigure}[b]{\linewidth}
		\centering
		\includegraphics[width=0.9\linewidth, trim={1cm 35 12 12}, clip]{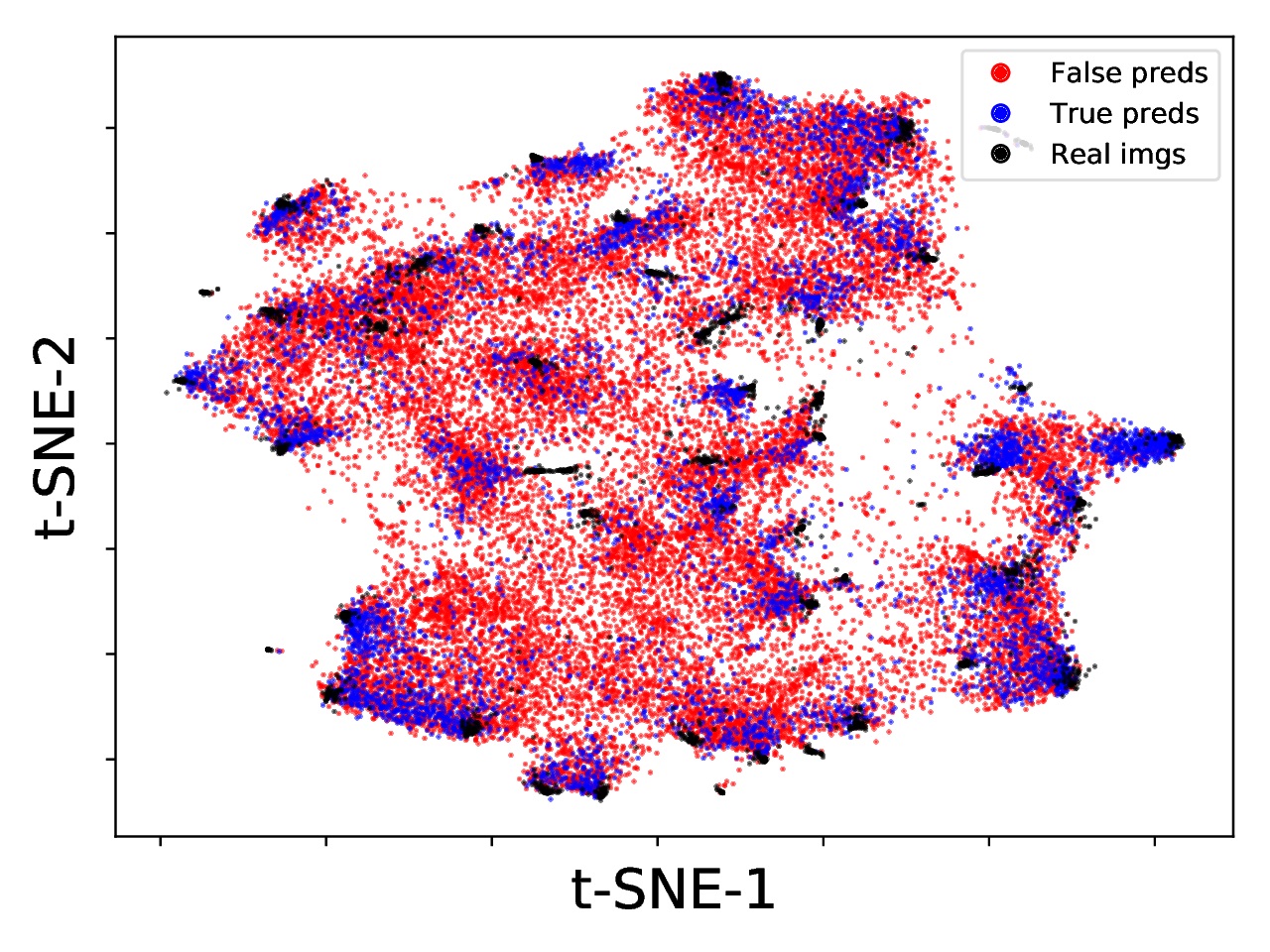}
		\caption{AttnGAN}
		\label{fig:attngan}
	\end{subfigure}%
	
	\begin{subfigure}[b]{\linewidth}
		\centering
		\includegraphics[width=0.9\linewidth, trim={1cm 35 12 12}, clip]{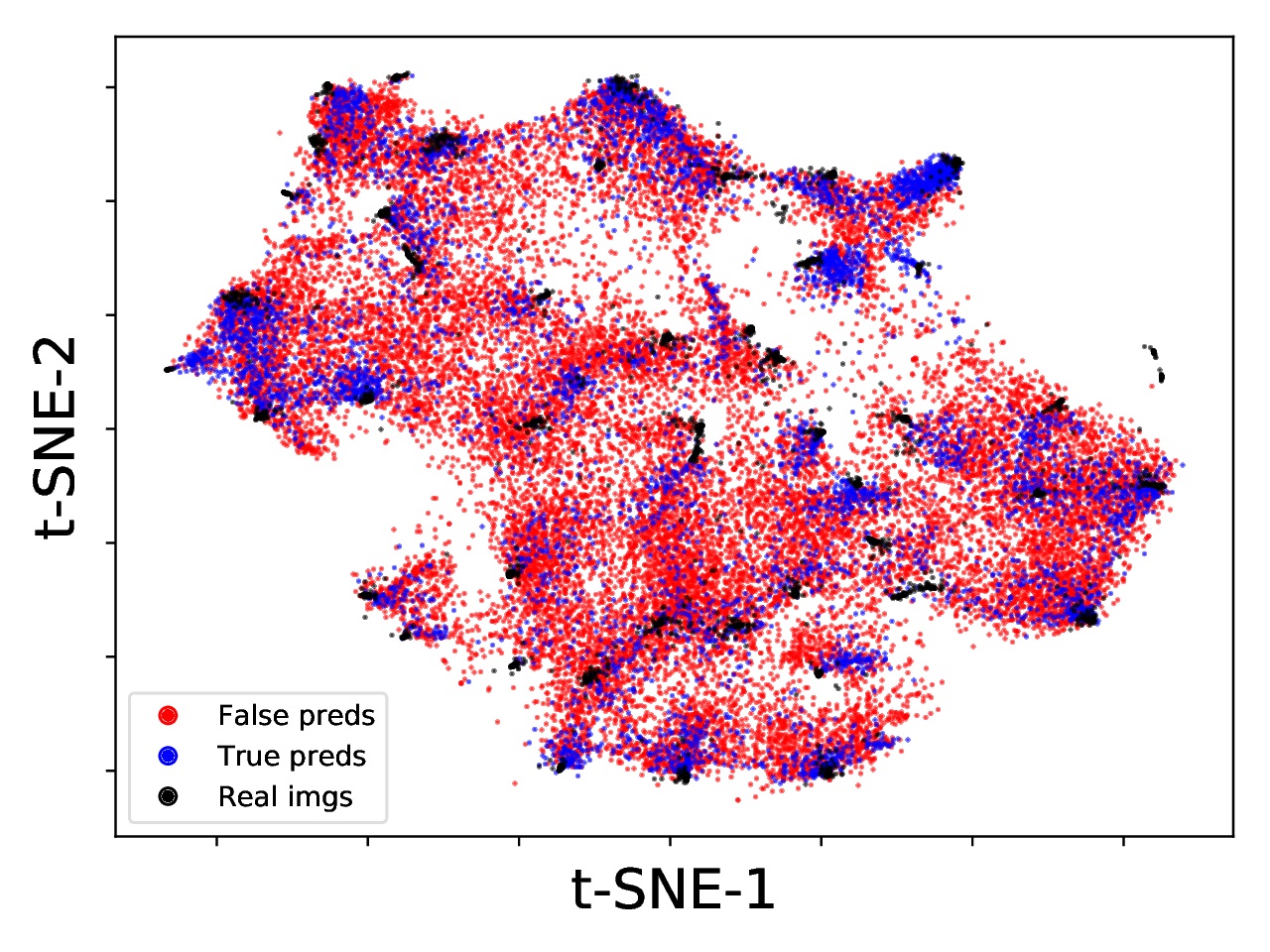}
		\caption{MirrorGAN}
		\label{fig:mirrorgan}
	\end{subfigure}
	\caption{$2$-dimensional image representations obtained by applying t-SNE on Inception features for images generated by (a) AttnGAN, and (b) MirrorGAN. Black points correspond to real images.}
	\label{fig:tsne1}
\end{figure}

\begin{figure}[ht]
	\centering
	\begin{subfigure}[b]{\linewidth}
		\centering
		\includegraphics[width=0.9\linewidth, trim={1cm 35 12 12}, clip]{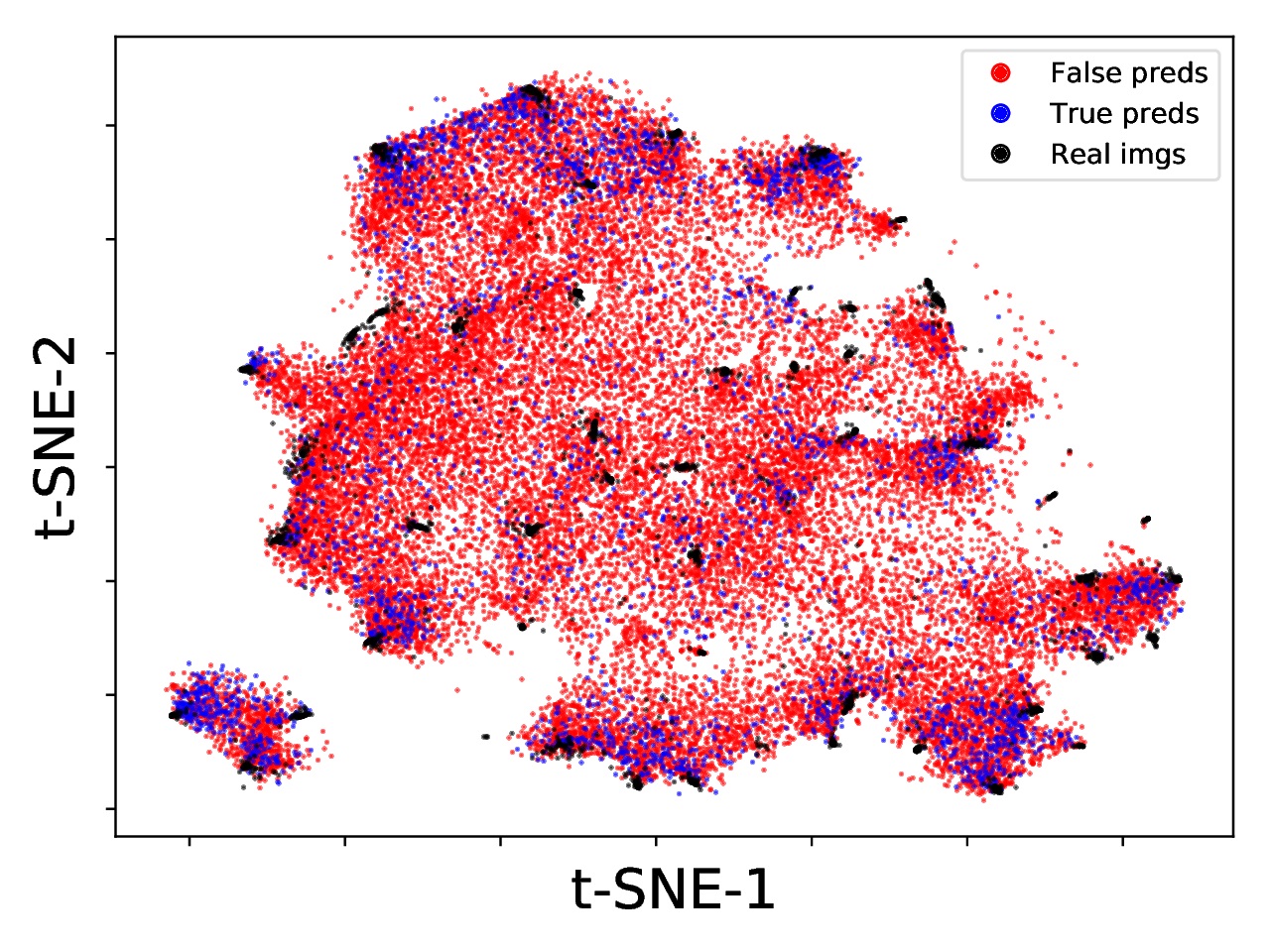}
		\caption{HDGAN}
		\label{fig:hdgan}
	\end{subfigure}%
	
	\begin{subfigure}[b]{\linewidth}
		\centering
		\includegraphics[width=0.9\linewidth, trim={1cm 35 12 12}, clip]{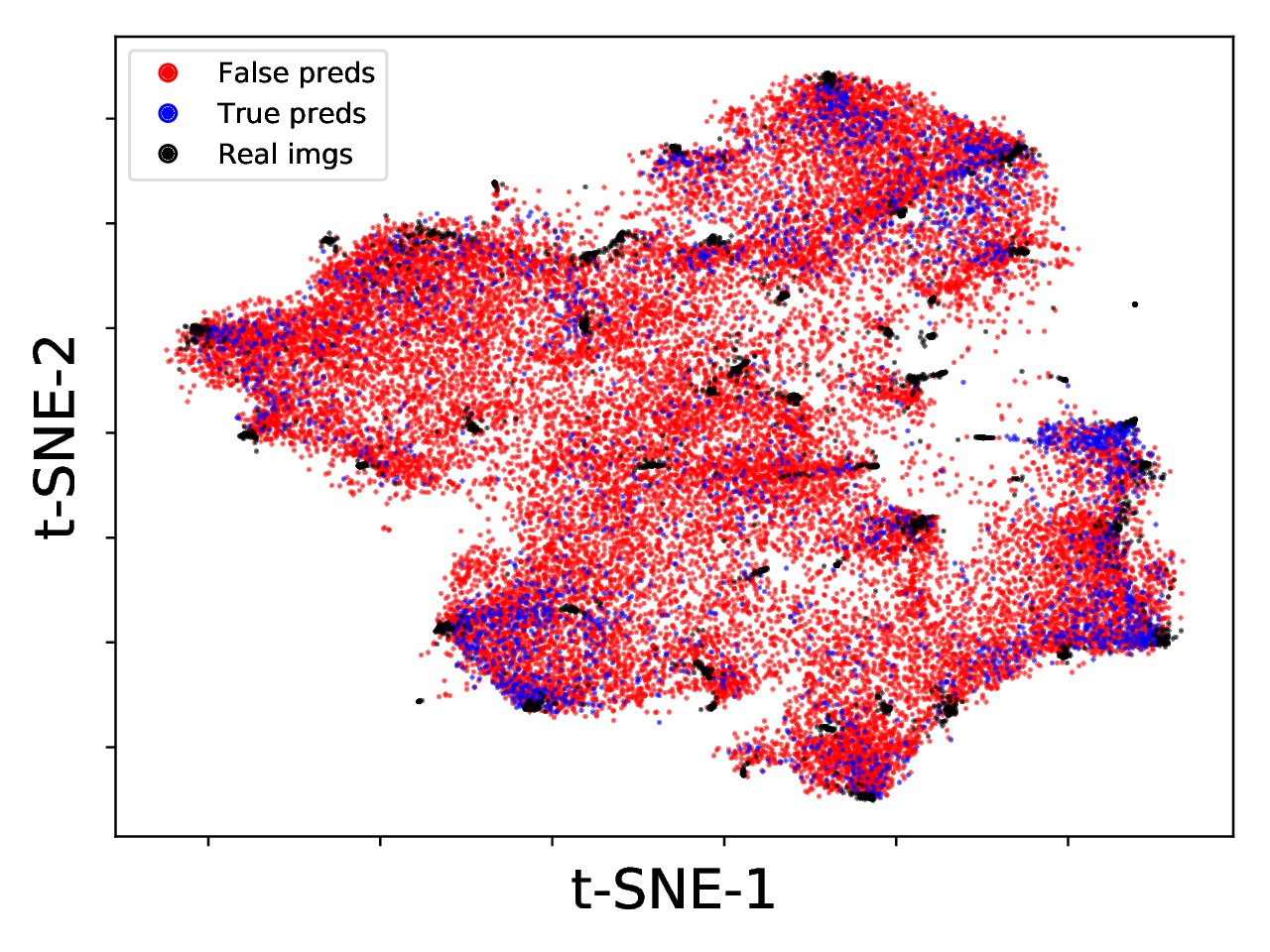}
		\caption{StackGAN}
		\label{fig:StackGAN}
	\end{subfigure}
	\caption{$2$-dimensional image representations obtained by applying t-SNE on Inception features for images generated by (a) HDGAN, and (b) StackGAN. Black points correspond to real images.}
	\label{fig:tsne2}
\end{figure}

For a properly trained Inception network, the image representations obtained by applying the t-SNE mapping of the real images $\Phi^X$ is expected to form tight clusters, corresponding (sub-)classes of the data set $(X,T)$. This is verified by our experiments, as can be seen from the $2$-dimensional plots in Figures \ref{fig:tsne1} and \ref{fig:tsne2}, where the real image representations illustrated as black points are well-clustered in disjoint groups. The distribution of the representations $\Phi^S$ of the synthetic images reveals the following properties:

\begin{description}
	\item \textbf{Realism:} For fairly realistically looking synthetic images, the distribution of their low-dimensional representations $\Phi^S$ should show a significant degree of overlap with the distribution of the low-dimensional representations of the real images $\Phi^X$. This is assuming the Inception network is not completely over-fitted to the real images (hence the choice of $\sim 95\%$ classification accuracy vs. $100\%$).
	
	\item \textbf{Variety:} The low-dimensional synthetic image representations $\Phi^S$ should be spread out to a reasonable degree, i.e. comparable to the distribution of low-dimensional representations of the real images $\Phi^X$. Too tight clustering of $\Phi^S$ signifies very similarly looking generated images or, in the worst case, mode collapse where the same image is generated from multiple different captions.
	
	\item \textbf{Semantic accuracy:} The types of data considered in this paper, i.e. CUB-200-2011 \cite{cub200} (see Figure \ref{fig:CUB_images}), consists of images belonging to distinct classes of similar looking objects, such as species, along with matching, human written captions. For such data, captions within each class should ideally show a high degree of semantic similarity, since they are all describing the same type of object (i.e. a specific species of bird). When the generator $G$ is successful at capturing the semantics of the captions in the set $T$, the generated images should not only be realistically looking but also similar to the real images belonging to the same class. This means that the distance between points in $\Phi^S$ and $\Phi^X$ belonging to the same classes should be a good gauge of the text-to-image synthesis method semantic accuracy.
	
\end{description}

To quantify the results of the process described above, a clustering scheme is used. First, fuzzy C-Means clustering \cite{fuzzycmeans} is applied on the real image representations $\Phi^X$. The number of cluster centers is set to the number of classes in the data set $(X,T)$. After clustering is performed, each cluster is assigned a label indicated by the most frequent class among the images belonging to that cluster. When two or more clusters are assigned the same class based on the majority of their images, the cluster containing most images of that class is given first priority when assigning labels. This ensures unique labels for the resulting clusters.

Each synthetic image $i$ represented by a low-dimensional vector in $\Phi^S$ is assigned probabilities $p_{ij}$ correspoding to the membership values with respect to the $j$\textsuperscript{th} cluster based on its distance to the corresponding cluster centroid. The predicted classes are then given by:
\begin{equation}
y^\mathrm{pred}_i=\underset{j}{\arg\max}\:\: p_{ij}.
\end{equation}
Since the true classes $y^\mathrm{true}_i$ of points in $\Phi^S$ are known, the clustering accuracy can be easily calculated by:	
\begin{align}
\mathrm{cluster\,acc} = \frac{1}{n}\sum_i
\begin{cases}
y^\mathrm{pred}_i = y^\mathrm{true}_i & 1\\
y^\mathrm{pred}_i \neq y^\mathrm{true}_i & 0.
\end{cases}
\label{eq:acc}
\end{align}

When $G$ correctly captures the semantics of the text descriptions in $T$ for a specific class, the resulting low-dimensional image representations should lie close to the cluster of real images belonging to the corresponding class. Thus, the clustering accuracy above should be a good metric for evaluating the ability of the text-to-image synthesis method to synthesize images capturing the class semantics.

\begin{table*}[ht]
	\centering
	\begin{tabular}{lllll} \toprule
		& AttnGAN & HDGAN & MirrorGAN & StackGAN-v2 \\ \midrule 
		Inception score & \multicolumn{1}{r}{\SI{4.32 \pm 0.05}} & \multicolumn{1}{r}{\SI{4.00 \pm 0.05}} & \multicolumn{1}{r}{\SI{4.53 \pm 0.04}} & \multicolumn{1}{r}{\SI{4.09 \pm 0.05}} \\
		Self-reported & \multicolumn{1}{r}{\SI{4.36\pm 0.03}} & \multicolumn{1}{r}{\SI{4.15\pm 0.05}} & \multicolumn{1}{r}{\SI{4.56 \pm 0.05}} & \multicolumn{1}{r}{\SI{4.04\pm 0.05}}  \\ \hline 
		
		
		Clustering acc. (real) & \multicolumn{1}{r}{\SI{0.96}} & \multicolumn{1}{r}{\SI{0.95}} & \multicolumn{1}{r}{\SI{0.95}} & \multicolumn{1}{r}{\SI{0.96}} \\
		
		
		\textbf{Clustering acc. (synthetic)} & \multicolumn{1}{r}{\textbf{0.22}} & \multicolumn{1}{r}{\textbf{0.11}} & \multicolumn{1}{r}{\textbf{0.20}} & \multicolumn{1}{r}{\textbf{0.11}}  \\ \hline
		
		Inception acc. (synthetic) & \multicolumn{1}{r}{\SI{0.23}} & \multicolumn{1}{r}{\SI{0.13}} & \multicolumn{1}{r}{\SI{.22}} & \multicolumn{1}{r}{\SI{0.12}} \\ \bottomrule
	\end{tabular}
	\caption{Inception scores and results of fuzzy c-means clustering on the t-SNE embeddings for images generated from the CUB-200-2011 test set. The errors given correspond to $1$ standard deviation.}
	\label{tab:evaluation_scores}
\end{table*}

\section{Experiments}
In this Section, we describe experiments conducted to evaluate the ability of the proposed method to capture semantic class information in text-to-image synthesis methods evaluation. We used four state-of-the-art text-to-image synthesis methods to this end, i.e. the \emph{AttnGAN} \cite{attngan}, the \emph{MirrorGAN} \cite{mirrorgan}, the \emph{HDGAN} \cite{hdgan}, and the \emph{StackGAN++} \cite{stackgan}. These methods were used to generate synthetic images using each of the $\sim 30$k captions from the test set of the CUB-200-2011 dataset \cite{cub200}. The captions describe $\sim 3$k images of birds from $50$ different species ($10$ captions per image).

The pre-trained Inception model from \cite{stackgan}, trained on the CUB-200-2011 test data \cite{cub200}, was then used to generate image representations for the real images as well as the resulting synthetic images obtained by each of the four text-to-image synthesis methods. When performing the t-SNE mapping, the perplexity was set to 600 since this is the approximate number of synthetic images per class (and thus the expected number of neighbors).

Figures \ref{fig:tsne1} and \ref{fig:tsne2} show the $2$-dimensional data representations obtained by applying t-SNE on the real images combined with each of the synthetic image sets of the studied methods. The synthetic image representations are shown in blue and red, while the real image representations are shown in black. The blue points are synthetic image mappings that have been classified correctly using the proposed method based on fuzzy C-means. These are positioned around real image clusters corresponding to the correct species. The red points are representations of synthetic images that were not classified correctly. More blue points signify that the corresponding text-to-image synthesis method is better at capturing the semantics of the captions.

The AttnGAN (Figure \ref{fig:attngan}) and MirrorGAN (Figure \ref{fig:mirrorgan}) seem to yield the most correct classifications. Their t-SNE plots are very similar showing a high degree of clustering around the real image representations. They both seem to have a large amount of variety, and the methods are able to generate correctly classified images for almost all classes. There exists, however, still a large number of incorrectly classified images that are not similar to any real image cluster.

The StackGAN++ (Figure \ref{fig:StackGAN}) and HDGAN (Figure \ref{fig:hdgan}) methods are also very similar to each other but synthesize much fewer correctly classified images. They both show an equal amount of variety, similar to the AttnGAN and MirrorGAN frameworks, and produce images similar to all real image clusters. However, most of these images are miss-classified, revealing a lower ability to accurately portray the semantics of the captions.

Calculating the clustering accuracy (Eq. (\ref{eq:acc})) confirms the conclusions from the visual inspection. In Table \ref{tab:evaluation_scores} the clustering accuracy of each method is highlighted in bold font. The AttnGAN and MirrorGAN methods achieve an accuracy of 0.22 and 0.20, respectively which is close to double the accuracy achieved by the HDGAN and StackGAN++ methods. As a sanity check, the classification accuracy of the pre-trained Inception network was calculated directly (bottom line of table \ref{tab:evaluation_scores}), and the results are almost completely equal to those reported by the corresponding works. This is a strong indicator that the overall structure and clustering of the data has not been lost after applying the the t-SNE mapping.

The Inception scores were also calculated for the generated images. The Inception scores obtained by our experiments and those reported in the papers proposing each corresponding method are listed in the top two lines of table \ref{tab:evaluation_scores}. As can be seen, the self-reported scores are close to those reported by the corresponding papers. HDGAN's Inception score deviates somewhat from the self-reported score (3 standard deviations) while the other scores agree with self-reported results.

\section{Conclusion}
In this paper, we introduced a novel evaluation method and metric for text-to-image synthesis methods. We showed that the proposed visualization method allows for the assessment of several key properties of the generated images, namely realism, variety (including the presence of mode-collapse) and semantic similarity to the input text descriptions. These properties can be assessed for thousands of images simultaneously and lead to much better insight of the text-to-image synthesis methods' performance than the traditional evaluation approach based on the Inception Score \cite{inceptionscore}. Finally, a semantic accuracy metric was introduced that allows for easy comparison between different text-to-image synthesis methods' ability to accurately capture semantic information.  

\bibliographystyle{IEEEbib}
\bibliography{refs}




%
\newpage
\onecolumn
\section*{\hfil SUPPLEMENTARY MATERIAL \hfil}
\section*{Appendix: Detailed results for t-SNE of Inception-net features.}
The following plots shows the distribution of classes (color coded by class) for both generated and ground truth image features as well as the fuzzy-C-means\cite{fuzzycmeans} clustering results. Results are shown for the AttnGAN\cite{attngan}, HDGAN\cite{hdgan}, MirrorGAN\cite{mirrorgan} and StackGAN++\cite{stackgan++} methods.

\begin{figure}[h]
	\centering
	\includegraphics[width=1\linewidth]{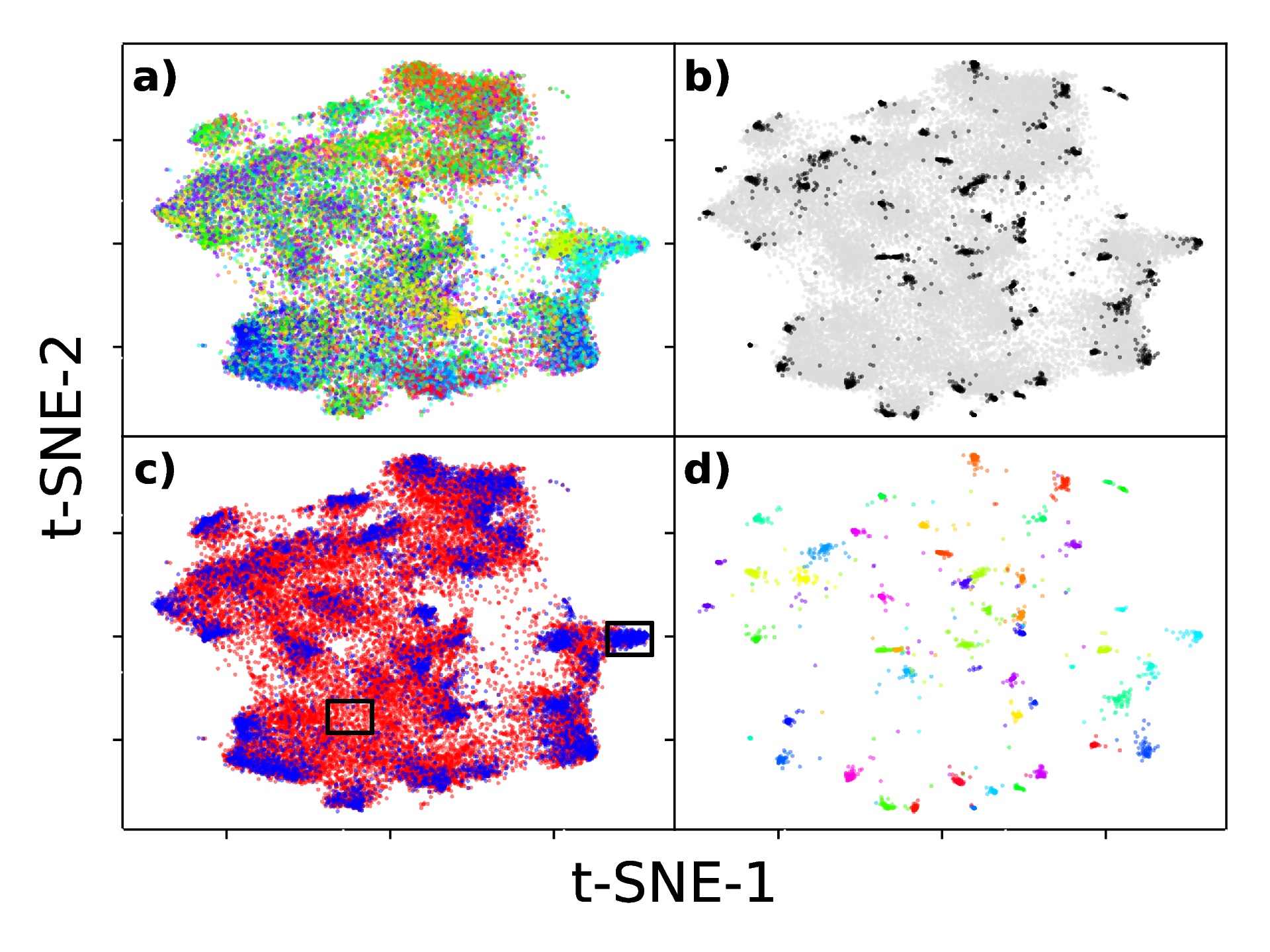}
	\caption{T-SNE results for \textbf{AttnGAN}\cite{attngan} on the CUB-200-2011\cite{cub200} test set (images generated from $\sim30$k captions). \textbf{a)} Generated image features colored by class. \textbf{b)} Real image features (black) and generated image features (gray). \textbf{c)} Results of fuzzy c-means clustering. Correct predictions are in blue, incorrect predictions are in red. Example images from the two black boxes are shown in figure \ref{fig:boxes}.  \textbf{d)} All $\sim 3$k real image features, colored by class.}
	\label{fig:tsne_attngan}
\end{figure}

\begin{figure}[h]
	\begin{subfigure}{.5\textwidth}
		\centering
		\includegraphics[width=0.95\linewidth]{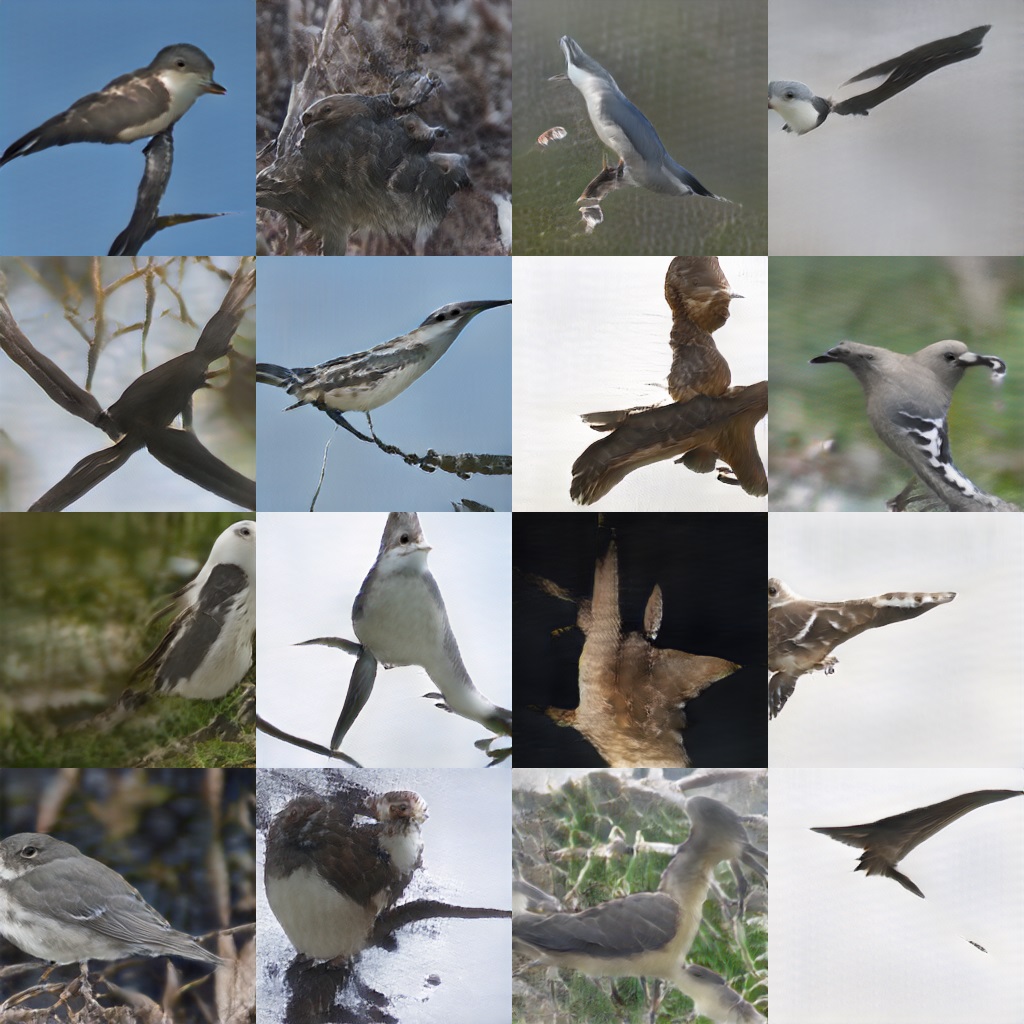}  
		\caption{}
	\end{subfigure}
	\begin{subfigure}{.5\textwidth}
		\centering
		\includegraphics[width=0.95\linewidth]{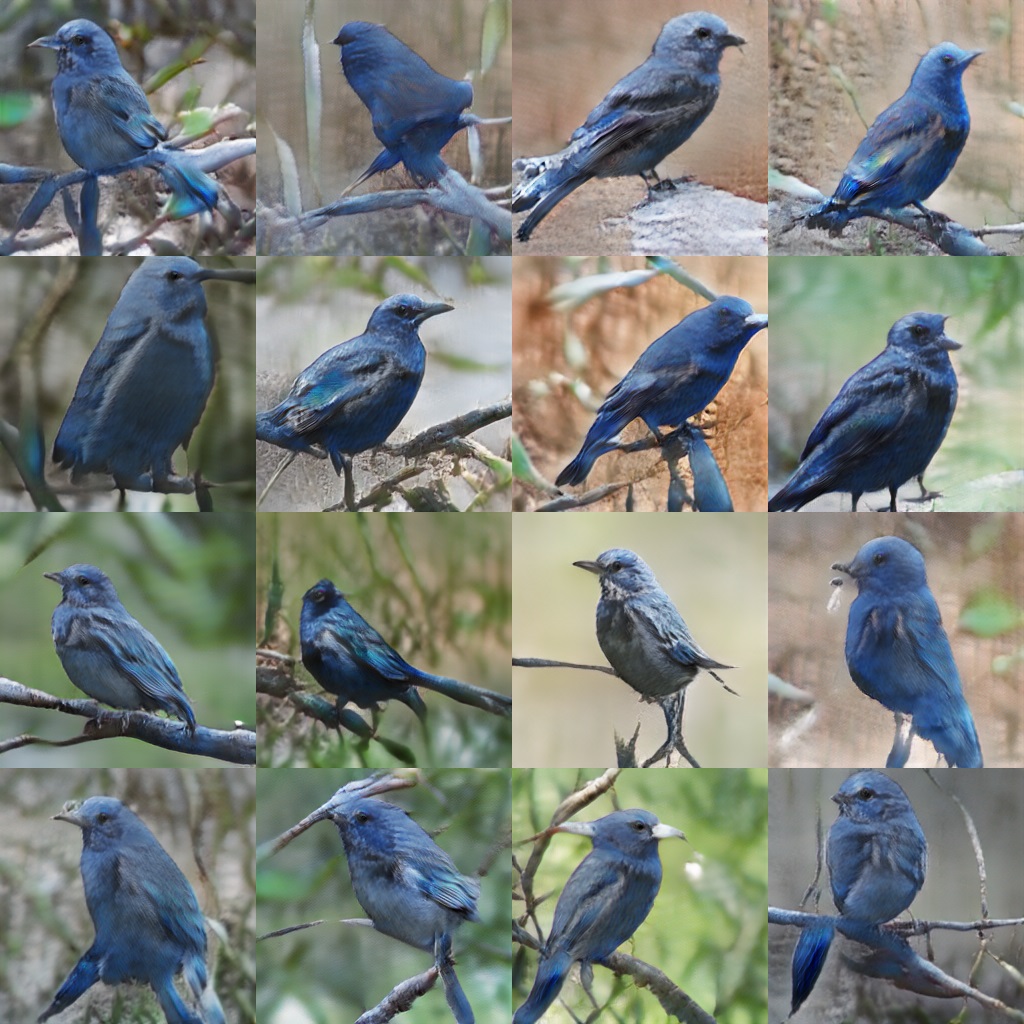}  
		\caption{}
	\end{subfigure}
	\caption{\textbf{a)} Random example images from the left black box in figure \ref{fig:tsne_attngan}c. \textbf{b)} Random example images from the right black box in figure \ref{fig:tsne_attngan}c. As can be seen, generated images falling far away from the representations of real images (and incorrectly classifier) illustrate a strange content. On the opposite, generated images falling close to the representations of real images (and correctly classifier) illustrate a visually plausible content.}
	\label{fig:boxes}
\end{figure}

\begin{figure}
	\centering
	\includegraphics[width=1\linewidth]{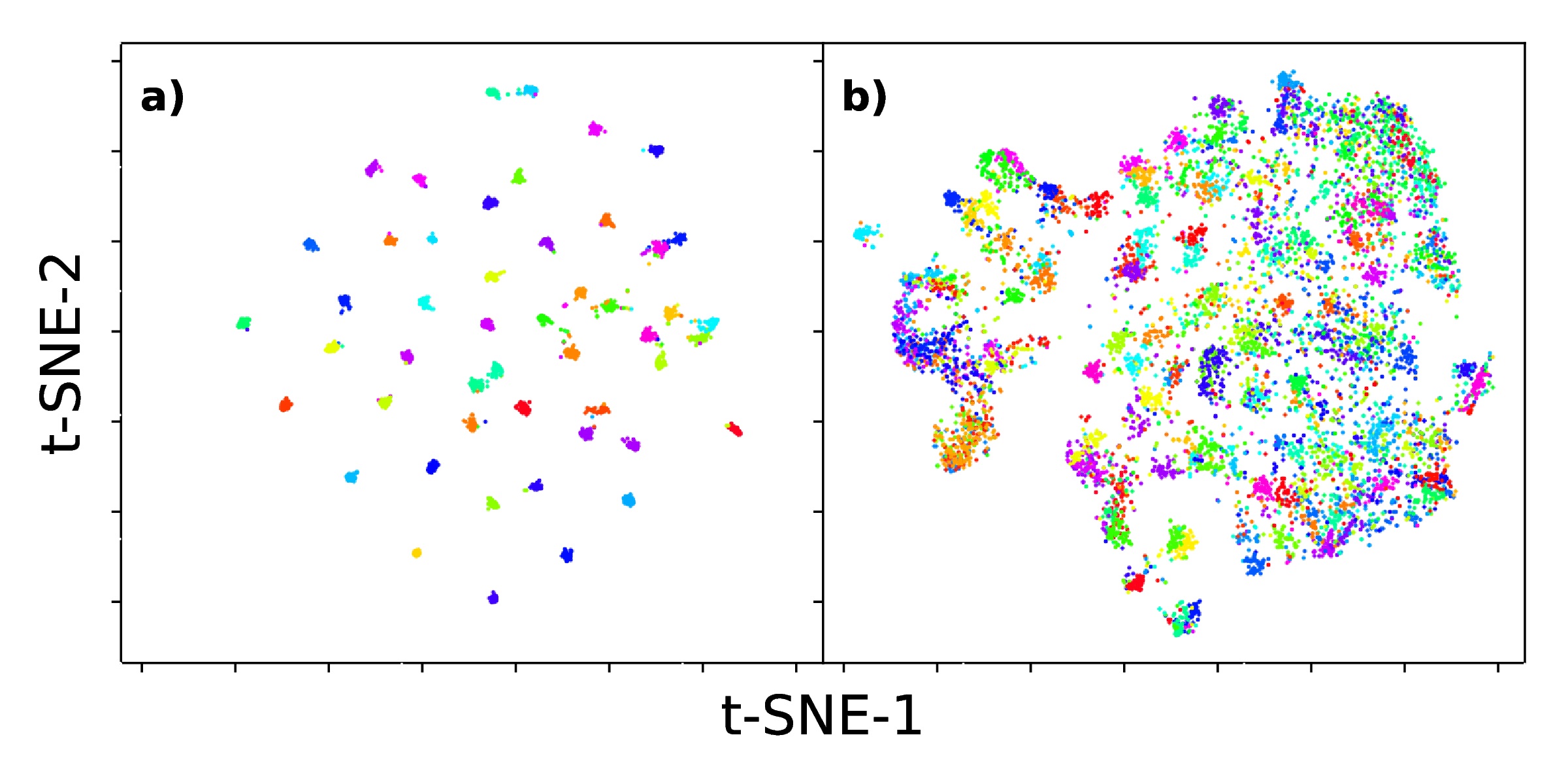}
	\caption{\textbf{a)} t-SNE mapping of Inception-net\cite{inceptionnet} features for the real images in the CUB-200-2011\cite{cub200} \textbf{test} set. \textbf{b)} t-SNE mapping of Inception-net features for the real images in the CUB-200-2011 \textbf{training} set. Points are colored by image class. Figure (a) clearly shows that the Inception-network is properly trained to classify the CUB-200-2011 test set since each class is tightly clustered. Figure (b) shows that the network is able to cluster images from unseen classes (the training set is disjoint from the test set) and thus able to generalize to new images.}
	\label{fig:tsne_train_test}
\end{figure}

\begin{figure}
	\centering
	\includegraphics[width=1\linewidth]{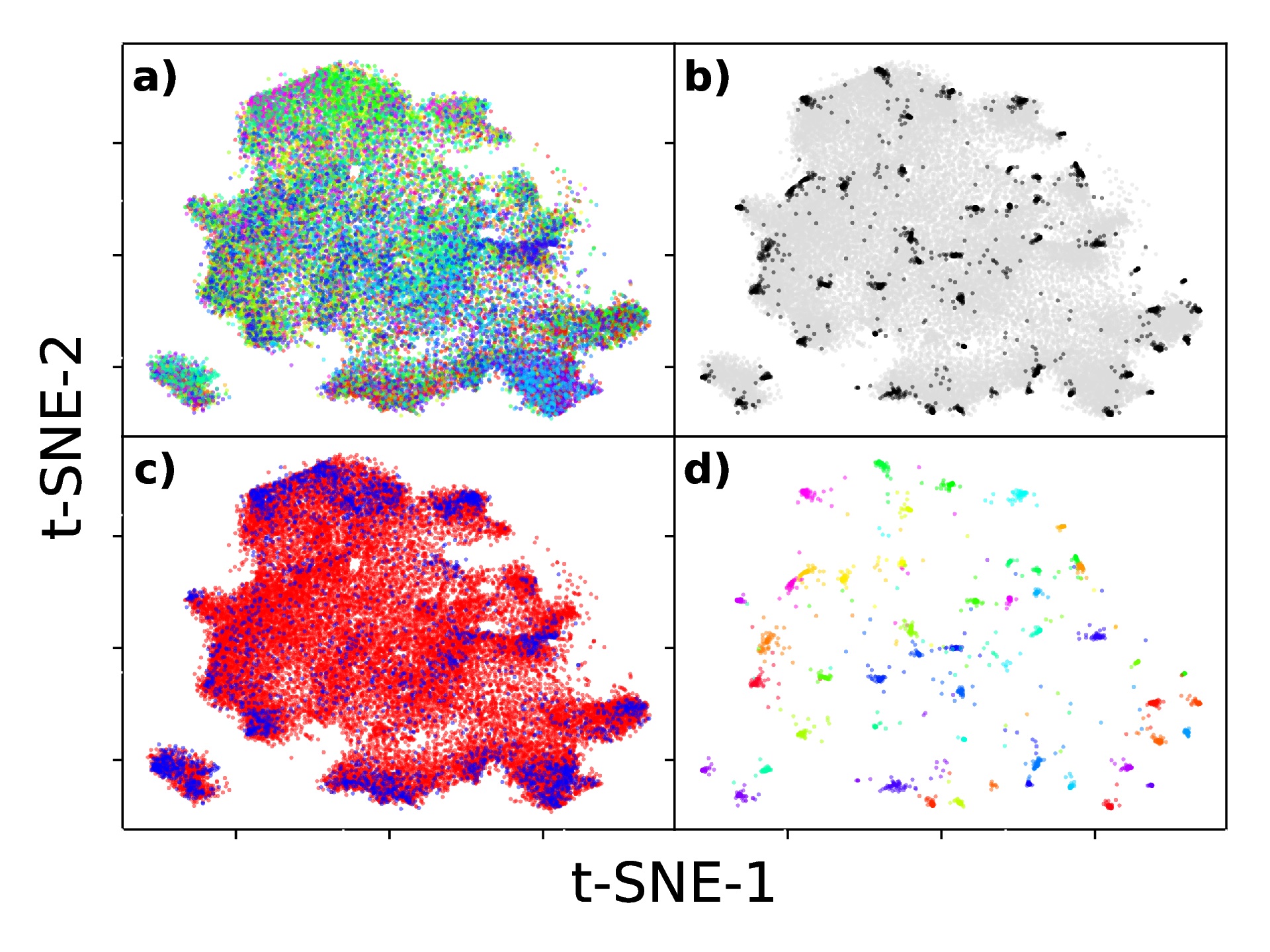}
	\caption{T-SNE results for \textbf{HDGAN}\cite{hdgan} on the CUB-200-2011\cite{cub200} test set (images generated from $\sim30$k captions). \textbf{a)} Generated image features colored by class. \textbf{b)} Real image features (black) and generated image features (gray). c) Results of fuzzy c-means clustering. Correct predictions are in blue, incorrect predictions are in red. \textbf{d)} All $\sim 3$k real image features, colored by class.}
	\label{fig:tsne_hdgan}
\end{figure}

\begin{figure}
	\centering
	\includegraphics[width=1\linewidth]{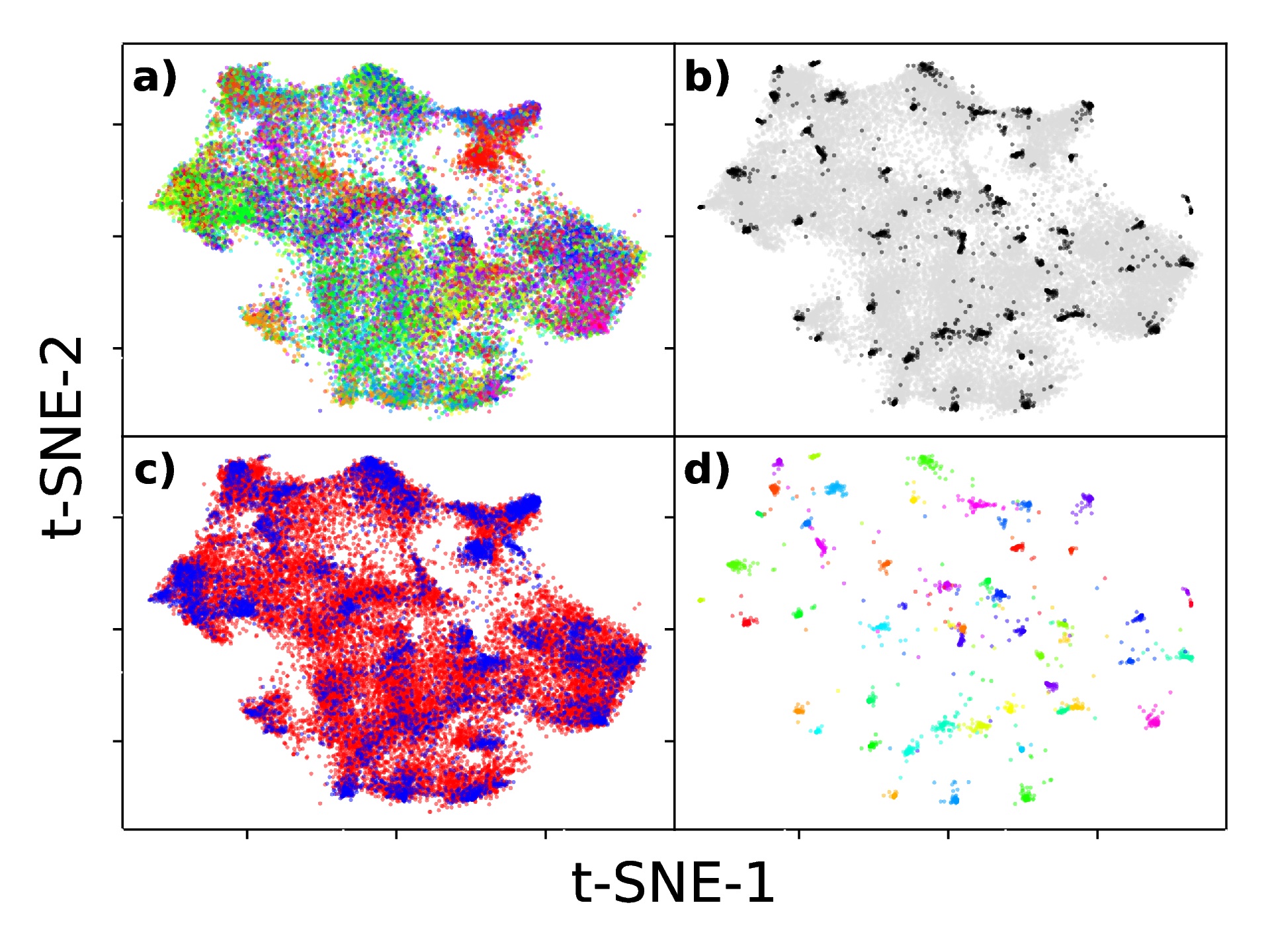}
	\caption{T-SNE results for \textbf{MirrorGAN}\cite{mirrorgan} on the CUB-200-2011\cite{cub200} test set (images generated from $\sim30$k captions). \textbf{a)} Generated image features colored by class. \textbf{b)} Real image features (black) and generated image features (gray). c) Results of fuzzy c-means clustering. Correct predictions are in blue, incorrect predictions are in red. \textbf{d)} All $\sim 3$k real image features, colored by class.}
	\label{fig:tsne_mirrorgan}
\end{figure}

\begin{figure}
	\centering
	\includegraphics[width=1\linewidth]{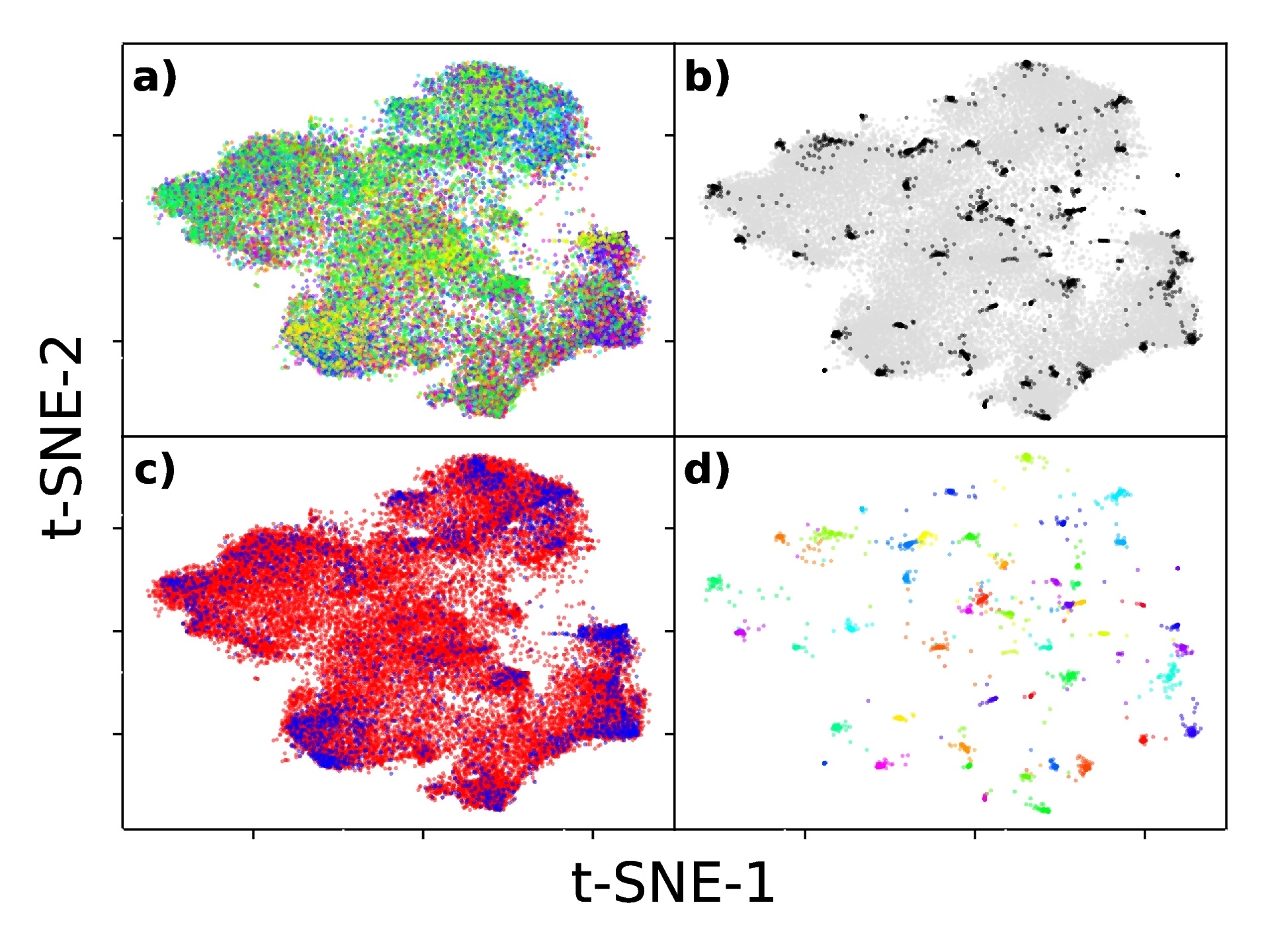}
	\caption{T-SNE results for \textbf{StackGAN++}\cite{stackgan++} on the CUB-200-2011\cite{cub200} test set (images generated from $\sim30$k captions). \textbf{a)} Generated image features colored by class. \textbf{b)} Real image features (black) and generated image features (gray). c) Results of fuzzy c-means clustering. Correct predictions are in blue, incorrect predictions are in red. \textbf{d)} All $\sim 3$k real image features, colored by class.}
	\label{fig:tsne_stackgan}
\end{figure}

\clearpage
\bibliographystyle{IEEEbib}
\bibliography{refs}

\end{document}